\documentclass{article}

\usepackage{hyperref}

\usepackage[accepted]{icml2018}

\usepackage{graphics}      

\usepackage{savesym}
\usepackage{amsmath}

\usepackage{color}

\usepackage{microtype}        
\usepackage{ccicons}          

\usepackage{relsize} 
\usepackage{mathtools,xparse}
\usepackage[table]{xcolor}
\usepackage{verbatim}
\usepackage{comment}
\usepackage{url}
\usepackage{soul}
\usepackage{multicol} 
\usepackage{multirow}
\usepackage{colortbl} 
\usepackage{amssymb}
\usepackage{wasysym}
\makeatletter
\def\oversortoftilde#1{\mathop{\vbox{\m@th\ialign{##\crcr\noalign{\kern3\p@}%
				\sortoftildefill\crcr\noalign{\kern3\p@\nointerlineskip}%
				$\hfil\displaystyle{#1}\hfil$\crcr}}}\limits}

\def\sortoftildefill{$\m@th \setbox\z@\hbox{$\braceld$}%
	\braceld\leaders\vrule \@height\ht\z@ \@depth\z@\hfill\braceru$}

\makeatother

\usepackage{tikz-cd}

\usepackage{tabulary}
\usepackage{booktabs}

\usepackage[english]{babel}
\usepackage{enumitem}

\newtheorem{theorem}{Theorem}
\newtheorem{prop}{Proposition}

\usepackage{array}
\usepackage{bytefield}

\usepackage[labelformat=simple,font={normal}, labelsep=period,singlelinecheck=on]{caption}
\usepackage[labelformat=simple,font={normal}, labelsep=period,singlelinecheck=on]{subcaption}

\usepackage{cleveref}
\crefname{section}{§}{§§}
\Crefname{section}{§}{§§}

\usepackage{slashbox}
\usepackage{multirow}
\usepackage{xcolor}

\usepackage{dsfont}
\usepackage{transparent}
\usepackage{import}
\usepackage{calrsfs}
\DeclareMathAlphabet{\pazocal}{OMS}{zplm}{m}{n}

\newcommand{\Sa}{\pazocal{S}}

\newcommand{\Ca}{\pazocal{C}}

\newcommand{\Na}{\pazocal{N}}
\newcommand{\BigO}{\pazocal{O}}
\usepackage{amsfonts}

\usepackage{tikz}
\definecolor{mygreen}{RGB}{0,176,80}
\definecolor{myred}{RGB}{102,0,0}
\definecolor{myblue}{RGB}{0,0,102}
\definecolor{myblue2}{RGB}{0,51,102}
\definecolor{myblue3}{RGB}{0,76,153}   
\definecolor{myblue4}{RGB}{48, 144, 199}  

\usetikzlibrary{decorations.pathreplacing}
\usetikzlibrary{bending}
\usepackage{lipsum}

\usetikzlibrary{positioning}
\tikzset{main node/.style={circle,fill=blue!20,draw,minimum size=1cm,inner sep=0pt},
}

\usepackage{etoolbox}
\patchcmd{\footnotemark}{\stepcounter{footnote}}{\refstepcounter{footnote}}{}{}

\usepackage{tabulary}
\newcolumntype{K}[1]{>{\centering\arraybackslash}p{#1}}
\usepackage{makecell}

\icmltitlerunning{Graph Capsule Convolutional Neural Networks }

\begin{document}

\twocolumn[
\icmltitle{Graph Capsule Convolutional Neural Networks }



\icmlsetsymbol{equal}{*}

\begin{icmlauthorlist}
\icmlauthor{Saurabh Verma}{ed}
\icmlauthor{Zhi-Li Zhang}{ed}
\end{icmlauthorlist}

\icmlaffiliation{ed}{Department of Computer Science, Univeristy of Minnesota Twin Cities USA}

\icmlcorrespondingauthor{Saurabh Verma}{verma076@cs.umn.edu}
\icmlcorrespondingauthor{Zhi-Li Zhang}{zhzhang@cs.umn.edu}

\icmlkeywords{Machine Learning, ICML}

\vskip 0.3in
]



\printAffiliationsAndNotice{}  

\begin{abstract}

Graph Convolutional Neural Networks (GCNNs) are the most recent exciting advancement in   deep learning field  and their applications are quickly spreading in  multi-cross-domains including  bioinformatics,  chemoinformatics,  social networks,  natural language processing and computer vision. In this paper, we expose and  tackle some of the basic weaknesses of a   GCNN model with a capsule idea presented in~\cite{hinton2011transforming} and propose our Graph Capsule Network (GCAPS-CNN) model. In addition, we design our GCAPS-CNN model to solve especially  graph classification problem which current GCNN models find challenging. Through extensive experiments, we show that our proposed Graph Capsule Network can significantly   outperforms both the existing state-of-art deep   learning methods and graph kernels   on   graph classification benchmark datasets.
\end{abstract}

\section{Introduction}
Graphs are one of the most fundamental structures that have been widely used for representing many types of data. Learning on graphs such as graph semi-supervised learning, graph classification or  graph evolution have found wide applications  in domains such as  bioinformatics,  chemoinformatics,  social networks, natural language processing and computer vision. With remarkable successes of deep learning approaches in image  classification and object recognition that attain ``superhuman'' performance, there has been a surge of research interests in generalizing convolutional neural networks (CNNs) to structures beyond regular grids, i.e., from 2D/3D images to arbitrary structures such as graphs~\cite{bruna2013spectral,henaff2015deep,defferrard2016convolutional,kipf2016semi}. These convolutional networks on graphs are now commonly known as Graph Convolutional Neural Networks (GCNNs). The principal idea behind graph convolution  has been derived from the  graph signal processing domain~\cite{shuman2013emerging}, which has since been extended in different ways for a variety of purposes~\cite{duvenaud2015convolutional, gilmer2017neural, kondor2018covariant}. 

In this paper, we  expose three major limitations of the standard GCNN model commonly used in existing deep learning approaches on graphs, especially when applied to the graph classification  problem, and explore ways to overcome these limitations. In particular, we propose a new model, referred to as {\em Graph Capsule Convolution Neural Networks (GCAPS-CNN)}. It is  inspired by the notion of {\em capsules} developed in~\cite{hinton2011transforming}: capsules are new types of neurons which encapsulate more information in a local pool operation (e.g., a convolution operation in a CNN)  by computing a small vector of highly informative  outputs rather than just taking a scalar output. Our {\em graph capsule} idea is quite general and can be employed in any version of GCNN model either design for solving  graph semi-supervised problem or doing sequence learning on graphs via Graph Convolution Recurrent Neural Network models (GCRNNs).

 The first limitation of the standard GCNN model  is due to the basic graph convolution operation which is defined -- in its purest form -- as the aggregation of node values in a local neighborhood corresponding to each feature (or channel). As such,  there is a potential loss of information associated with  the basic graph convolution operation.  
This problem has been noted  before~\cite{hinton2011transforming}, but has not attracted much attention  until recently~\cite{sabour2017dynamic}. To address this limitation, we  propose to improve upon the basic graph convolution operation by introducing the notion of {\em graph capsules} which encapsulate more information about nodes in a local neighborhood, where the local neighborhood is defined in the same way as in the standard GCCN model. Similar to the original capsule idea proposed in~\cite{hinton2011transforming}, this is achieved by replacing the scalar output  of a graph convolution operation  with a small vector output containing higher order statistical information per feature. Another source of inspiration for our proposed GCAPS-CNN model comes from one of the most successful graph kernels -- the Weisfeiler-Lehman (WL)- subtree graph kernel~\cite{shervashidze2011weisfeiler} designed specifically for solving the graph classification problem.
 In WL-subtree graph kernel, node labels  (features) are collected from neighbors of each node in a local neighborhood  and  compressed injectively to form a new node label in each iteration. The histogram of these new node labels are concatenated in each iteration to serve  as a  graph invariant feature vector. The important point to notice here  is that  due to the injection process, one can recover  the exact node labels of local neighbors in each iteration without losing  track of them. In contrast, this is not possible in the standard GCNN model as the   input feature values of node neighbors are lost  after the graph convolution operation.

The second major limitation of the standard GCNN model is specific to its (in)ability in tackling the graph classification problem.  GCNN models cannot be applied   directly   because they are equivariant  ({\em not invariant}) with respect to the node order in a graph. To be precise, consider a graph $G$  with Laplacian $\mathbf{L}\in \mathbb{R}^{N \times N}$    and node feature matrix $\mathbf{X}\in \mathbb{R}^{N \times d}$. Let $f(\mathbf{X}, \mathbf{L}) \in \mathbb{R}^{N \times h} $ be the output function of a GCNN model where $N, d, h$ are the number of nodes, input dimension and  hidden dimension of node features, respectively.  Then,
\textit{$f(\mathbf{X}, \mathbf{L})$ is a permutation equivariant function, i.e., for any $\mathbf{P}$   permutation matrix $f(\mathbf{P}\mathbf{X}, \mathbf{PLP^{T}}) = \mathbf{P}f(\mathbf{X}, \mathbf{L})$. }
This specific permutation equivariance property prevent us from directly applying GCNN to a graph classification problem, since it cannot provide any guarantee that the outputs of  any two isomorphic graphs are always the same. Consequently,  a GCNN architecture needs an additional graph permutation invariant layer in order to perform the  graph classification task successfully. This invariant layer  also   needs to be differentiable  for end-to-end learning.

Very limited amount of efforts has been devoted to carefully designing such an invariant GCNN model for the purpose of graph classification. Currently the most common  method for achieving graph permutation invariance  is performing aggregation (i.e., summing) over all graph node values~\cite{atwood2016diffusion, dai2016discriminative, zhao2018substructure, simonovsky2017dynamic}. Though  simple and fast, it can again incur significant loss of information.  Likewise, using a max-pooling layer to achieve graph permutation invariance encounters similar issues. A few attempts have been made~\cite{zhang2018end, kondor2018covariant} that go beyond aggregation or max-pooling in designing graph permutation invariant GCNNs. In~\cite{zhang2018end} the authors  propose a global ordering of nodes by sorting them according to their   values in the last hidden layer. This type of invariance is based on  creating an order among nodes and has also been explored before in~\cite{niepert2016learning}. However, as discussed in Section~\ref{sec:max-issue}, we show that there  are some issues with this type of approach. A more tangential approach  has been adopted in~\cite{kondor2018covariant} based on group theory to  design transformation operations and tensor aggregation rules that results in permutation invariant outputs. However, this approach relies on computing high order tensors which are computationally  expensive in many cases. 
To that end, we propose a novel permutation invariant layer based on computing the covariance of the data whose output does not depend upon the order of nodes in the graph. It is also fast to compute since it requires only a single dense-matrix multiplication operation.

Our last concern with the standard GCNN model  is their limited ability in exploiting global information for the purpose of  graph classification. The  filters employed in graph convolutions  are in essence  local in nature and hence can only provide an ``average/aggregate view'' of the local    data. This shortcoming poses a serious difficulty in handling graphs  where node labels are not present; approaches which initialize (node) feature values using, e.g.,  node degree, are not much helpful in this respect.   We propose to utilize global features (features that account for the full graph structure)  using a family of graph spectral distances as proposed in~\cite{verma2017hunt} to remedy this problem.

\textbf{In summary, the major contributions of our paper are: }
 
\begin{itemize}[leftmargin=*]
	\setlength\itemsep{-0.1em}
\item  We propose a novel Graph Capsule Convolution Neural  Network  model based on the capsule idea to capture highly informative output in a small vector  in place of a scaler output currently employed in  GCNN models. 
\item  We develop a novel graph permutation invariant layer based on computing the covariance of data to solve graph classification problem. We show that it is a better choice than performing node aggregation or doing max pooling and at the same time it can be computed efficiently.
\item  Lastly, we  advocate explicitly including global graph structure features  at each graph node to enable the proposed GCAPS-CNN model to exploit them for graph learning tasks. 
\end{itemize}

We organize our paper into five sections. We start with the related work on graph kernels and GCNNs in Section~\ref{sec:related_work}, and present our core idea behind graph capsules  in Section~\ref{sec:model}. In Section~\ref{sec:permutation_inv}, we focus on building a graph permutation invariant layer especially for solving the graph classification problem. In Section~\ref{sec:global_features}, we propose to equip our GCAPS-CNN model with enhanced global features to   exploit the full  graph   structure for learning on graphs. Lastly in Section~\ref{sec:exp_results} we conduct experiments and show the superior performance of our proposed GCAPS-CNN model.

\section{Related Work}~\label{sec:related_work}

 
There are three main approaches for solving the graph classification problem. The most common approach is concerned with building graph kernels. In graph kernels, a graph $G$ is decomposed into (possibly different) $\{G_{s}\}$ sub-structures. The graph kernel $K(G_1,G_2)$   is defined based on the frequency of each sub-structure appeared in $G_1$ and $G_2$, respectively. Namely, $K(G_1,G_2)=\langle f_{G_{s_1}},f_{G_{s_2}}\rangle$, where $f_{G_{s}}$ is the vector containing  frequencies of  $\{G_{s}\}$ sub-structures, and $\langle,\rangle$ is an inner product in an appropriately defined normed vector space. Much of work has been devoted to deciding on which sub-structures are more suitable than  others. 
Among the existing graph kernels, popular ones are  graphlets~\cite{prvzulj2007biological,shervashidze2009efficient}, random walk and shortest path kernels~\cite{kashima2003marginalized,borgwardt2005shortest}, and Weisfeiler-Lehman subtree kernel~\cite{shervashidze2011weisfeiler}. Furthermore, deep graph kernels~\cite{yanardag2015deep}, graph invariant  kernels~\cite{orsini2015graph}, optimal  assignment graph kernels~\cite{kriege2016valid}  and  multiscale laplacian graph kernel~\cite{kondor2016multiscale}  have been proposed with the goal to re-define kernel functions   to appropriately capture sub-structural similarity at different levels.  Another line of research in this area focuses on  efficiently computing  these kernels either through exploiting certain structure dependency, or via approximation or randomization~\cite{feragen2013scalable,de2013fast,neumann2012efficient}.  



The second category involves constructing explicit graph features such as  \textsc{Fgsd} features in~\cite{verma2017hunt} which is based on a family of graph spectral distances. It comes with certain theoretical guarantees. The Skew Spectrum of Graphs~\cite{kondor2008skew} based on group-theoretic approaches  is another  example in this category. Graphlet spectrum~\cite{kondor2009graphlet}  improves upon this work by including   labeled information; it also accounts for the relative position of subgraphs  within a graph. However, the main concern with   graphlet spectrum or  skew spectrum is its computational $\BigO(N^{3})$ complexity.

The third -- more recent and perhaps more promising -- approach to the graph classification is on developing convolutional neural networks (CNNs) for graphs.  The original idea of defining graph convolution operations   comes from the graph signal processing domain~\cite{shuman2013emerging}, which  has since been recognized as the problem of learning   filter parameters that appear in the graph fourier transform in the form of a   graph Laplacian~\cite{bruna2013spectral,henaff2015deep}.  Various GCNN models such a~\cite{kipf2016semi,atwood2016diffusion, duvenaud2015convolutional} have been proposed, 
  where traditional graph filters are replaced by a self-loop graph adjacency matrix and the outputs of  each neural network layer output are computed using a propagation rule while updating the network weights.  The authors in~\cite{defferrard2016convolutional} extend such GCNN models by utilizing fast localized spectral filters and efficient  pooling operations. A very different approach is  proposed in~\cite{niepert2016learning}  where a set of  local nodes are converted into a  sequence in order to  create receptive fields which are then   fed into a 1D convolutional neural network.

Another popular name for GCNNs is message passing neural networks (MPNNs)~\cite{lei2017deriving, gilmer2017neural,dai2016discriminative, garcia2017learning} . Though the authors in~\cite{gilmer2017neural} suggests that GCNNs are a special case  of MPNNs, we believe that both are equivalent models in a certain sense; it is simply a matter of  how the graph convolution operation is defined. In MPNNs the  hidden states of each node  is updated based on messages received from its neighbors as well as  the values of the previous hidden states in each iteration.  
This is made possible    by replacing traditional neural networks in GCNN with a small recurrent neural network (RNN) with the same weight parameters shared across all nodes in the graph. Note that here the number of iterations in MPNNs can be related to the depth of a GCNN model. In~\cite{simonovsky2017dynamic} the  authors propose to condition the learning parameters of filters  based on edges rather than on traditional nodes. This approach is similar to some instances of MPNNs such as in ~\cite{gilmer2017neural} where learning parameters are also associated with  edges. All the above MPNNs models employ aggregation   as the graph permutation invariant layer for solving the graph classification problem. In contrast, the authors  in~\cite{zhang2018end, kondor2018covariant} employs a max-sort pooling layer and group theory to achieve graph permutation invariance.

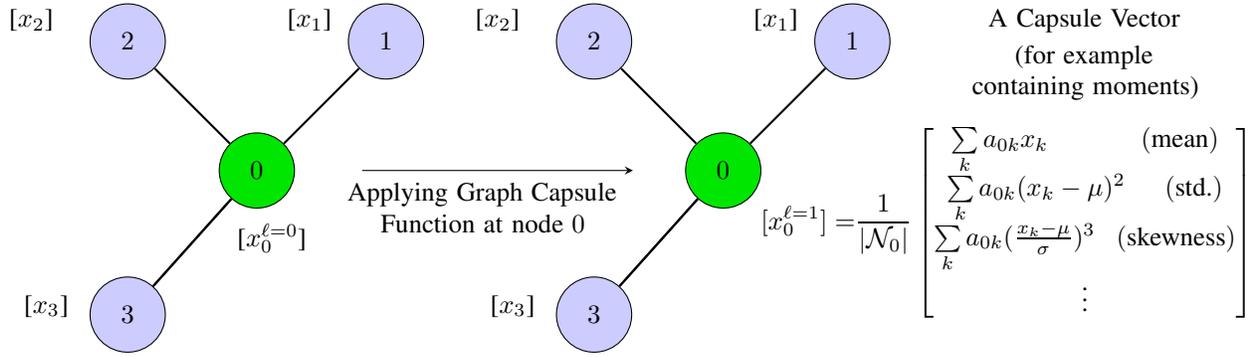
\begin{figure*}[!ht]
 	\centering
 	\vspace{1.5em}
\begin{tikzpicture}

\begin{scope}[xshift=-15cm]
\begin{scope}[xshift=-5cm]
\node[main node, fill=black!10!green] (0) {$0$};
\node[main node] (1) [above right = 1.0cm and 1.0cm of 0]  {$1$};
\node[main node] (2) [above left = 1.0cm and 1.0cm of 0] {$2$};
\node[main node] (3) [below left = 1.2cm and 1.0cm of 0] {$3$};

\path[draw,thick]
(0) edge node {} (1)
(2) edge node {} (0)
(0) edge node {} (3)
(3) edge node {} (0);

\node[] at (0.2, -0.9)   (a) {[$x^{\ell=0}_0$]};
\node[] at (-3, 2)   (a) {[$x_2$]};
\node[] at (0.7, 2)   (a) {[$x_1$]};
\node[] at (-2.8, -1.8)   (a) {[$x_3$]};

\end{scope}

\begin{scope}[xshift=1.2cm]
\node[main node, fill=black!10!green] (0) {$0$};
\node[main node] (1) [above right = 1.0cm and 1.0cm of 0]  {$1$};
\node[main node] (2) [above left = 1.0cm and 1.0cm of 0] {$2$};
\node[main node] (3) [below left = 1.2cm and 1.0cm of 0] {$3$};

\path[draw,thick]
(0) edge node {} (1)
(2) edge node {} (0)
(0) edge node {} (3)
(3) edge node {} (0);

\node[] at (-3, 2)   (a) {[$x_2$]};
\node[] at (0.7, 2)   (a) {[$x_1$]};
\node[] at (-2.8, -1.8)   (a) {[$x_3$]};
\end{scope}
\draw [->,>=stealth, scale=3] (-1.2,0) -- (0.0,.0);
\node[text width=4cm,align=center] at (-2, -0.5)   (a) {Applying Graph Capsule Function at node $0$};
\node[text width=4cm,align=center] at (6, 2)   (a) {A Capsule Vector};
\node[text width=4cm,align=center] at (6, 1.3)   (a) {(for example  \\ containing moments)};
\node[text width=4cm,align=center] at (3.7, -0.5)   (a) {\begin{equation*} 
	\begin{split}
	[x^{\ell=1}_0]= &    \frac{1}{|\Na_0|} \begin{bmatrix}
	\sum\limits_{k} a_{0k} x_k  \hspace{3.5em}(\text{mean} )\\
	\sum\limits_{k}a_{0k}( x_k-\mu)^2  \hspace{1.5em} (\text{std.} )  \\
	\sum\limits_{k}a_{0k} (\frac{x_k-\mu}{\sigma})^3   \hspace{0.8em} (\text{skewness} ) \\
	\vdots \\
	\end{bmatrix}\\
	\end{split}
	\end{equation*}};

\end{scope}

\end{tikzpicture}
\caption{Above figure shows that the  graph  capsule function at node $0$  computes a capsule vector which encodes higher-order statistical information about its local neighboorhood (per  feature). Here $\{x_0, x_1, x_2, x_3\}$ are respective node feature values. For example, when a node has no more than two neighbors then it is possible to recover back the input node neighbors   values from the very first three statistical moments.}
\label{fig:capsule}
\end{figure*}
 
\section{Graph Capsule CNN Model}\label{sec:model}

\noindent \textbf{Basic Setup and Notations}: Consider a   graph  $G=(V,E,\mathbf{A})$ of size $N=|V|$, where $V$ is  the vertex set, $E$ the edge set (with no self-loops) and $\mathbf{A}=[a_{ij}]$ the   weighted   adjacency matrix. The standard graph Laplacian is defined as $\mathbf{L}=\mathbf{D}-\mathbf{A} \in \mathbb{R}^{N \times N}$, where $\mathbf{D}$ is the degree matrix. Let $\mathbf{X}\in \mathbb{R}^{N \times d}$ be the node feature matrix, where $d$ is the input dimension.  When used, we will use $h$ to  denote the dimension of  hidden (latent) variables/feature space.

\noindent \textbf{General GCNN Model}: We start by describing a general GCNN model before presenting our Graph Capsule CNN model. Let  $G$ be a graph with graph  Laplacian $\mathbf{L}$  and $\mathbf{X}\in \mathbb{R}^{N \times d}$ be a node feature matrix.  Then the most general form of a GCNN layer output function $f(\mathbf{X}, \mathbf{L}) \in \mathbb{R}^{N \times h} $ equipped with polynomial filters  is given by Equation~(\ref{eq:gcnn}),


\vspace{-1em}
\begin{equation}\label{eq:gcnn}
\begin{split}
f(\mathbf{X}, \mathbf{L}) &  = \sigma\Bigg( 
\underbrace{ \begin{bmatrix} \mathbf{X} & \mathbf{L} \mathbf{X} & \dots &\mathbf{L}^k \mathbf{X} \end{bmatrix}}_{{g(\mathbf{X}, \mathbf{L})}} \hspace{-2em}  \underbrace{  \begin{bmatrix}
	\mathbf{W}_1   \\
	\mathbf{W}_2   \\
	\vdots \\
	\mathbf{W}_k   \\
	\end{bmatrix}}_{\text{learning weight parameters}}\hspace{-2em}  \Bigg)\\  
& = \sigma\Big(\sum_{k=0}^{K} \mathbf{L}^{k}\mathbf{X}\mathbf{W}_k \Big)\\
\end{split}
\end{equation}


In Equation~(\ref{eq:gcnn}),  $g(\mathbf{X},\mathbf{L}) \in \mathbb{R}^{N \times k   d} $ is defined as a graph convolution filter of polynomial form with degree $k$. While $[\mathbf{W}_1, \mathbf{W}_2,  ..., \mathbf{W}_k]$ are learning weight parameters where each $\mathbf{W}_k \in \mathbb{R}^{d \times h}$. 

Note that $g(\mathbf{X}, \mathbf{L})=[\mathbf{X}, \mathbf{L}\mathbf{X},..., \mathbf{L}^{K}\mathbf{X}] \in \mathbb{R}^{N \times k  d}$  can be seen as a new node feature matrix with extended dimension $kd$\footnotemark\label{fnm:1}\footnotetext{Also referred  to as  the breadth of a GCNN layer \label{fnt:1}.}. Furthermore, $\mathbf{L}$  can be replaced by any other suitable filter matrix as discussed in~\cite{levie2017cayleynets, kipf2016semi}.

A GCNN model with  a depth of $L$  layers can  be  expressed recursively as, 

\begin{equation}\label{eq:layer_gcnn}
\begin{split}
f^{(\ell)}(\mathbf{X}, \mathbf{L}) & = \sigma \big( g(f^{(\ell-1)}(\mathbf{X}, \mathbf{L}), \mathbf{L}) \mathbf{W}^{(\ell)}\big) \\
\end{split}
\end{equation}

where $\mathbf{W}^{\ell} \in \mathbb{R}^{kd \times h}$ is the  weight parameter matrix for the $\ell^{th}-$layer, $1 \leq l \leq L$.

One can notice that in any layer the basic computation expression involve is $[\mathbf{L}^{k}f^{(\ell-1)}(\mathbf{X}, \mathbf{L})]_{ij}$. This expression represents that the new $j^{th}$ feature value of  $i^{th}$ node  (associated   with the $i^{th}$ row)  is yielded out as a single (scalar) aggregated value based on its local-hood neighbors. This particular operation can   incur significant loss of information. We aim to remedy this issue by introducing our novel GCAPS-CNN model based on   the fundamental capsule idea.

\subsection{Graph Capsule Networks}\label{sec:graph_capsule}

The core idea behind our proposed graph capsule convolutional neural network is to   capture more information in a local node pool  beyond what is captured by aggregation, the graph convolution operation used in a standard GCCN model. This new information is encapsulated in so-called  instantiation parameters described in~\cite{hinton2011transforming} which forms a capsule vector of highly informative outputs. 

The quality of these  parameters are determined by their ability to encode the node feature values in a local neighborhood of each node  as well decode (i.e., to reconstruct) them from the capsule vector. For instance, one can take the histogram of neighborhood feature values as the  capsule vector. If histogram bandwidth is sufficiently small, we can guarantee to recover back  all the original input node   values. This strategy has been used  in constructing a successful graph kernel. However, as histogram is not a continuous differentiable function, it cannot be employed in backpropagation for  end-to-end deep learning.

Beside seeking representative instantiation parameters, we further impose two  more constraints on a graph capsule function. First, we  want our graph capsule function to be  permutation invariant (unlike equivariant as discussed in~~\cite{hinton2011transforming})  with respect to the input node order since  we are interested in a model that can produce the same output for isomorphic graphs.  Second,  we would like   to be able to compute these parameters efficiently.

\noindent \textbf{Graph Capsule Function}: To describe a general graph capsule function, consider an $i^{th}$ node  with $x_0$ value and the set of its neighborhood node values as $\Na(i)=\{x_0, x_1, x_2,...,x_k\}$ including itself. In the standard graph convolution operation, the output is a scalar function $f:\mathbb{R}^{k}\rightarrow \mathbb{R}$ which takes $k$  input neighbors at the $i^{th}$ node and yields an output given by

\begin{equation}\label{eq:gcnn_op}
\begin{split}
f_i(x_0, x_1,...,x_k) = \frac{1}{|\Na(i)|} \sum_{k \in \Na(i)} a_{ik} x_k \\
\end{split}
\end{equation}
where $a_{ik}$ represents   edge weights between nodes $i$ and $k$. 

In our graph capsule network, we replace $f(x_0,..., x_k)$ with a vector-valued capsule function $f:\mathbb{R}^{k}\rightarrow \mathbb{R}^{p}$. For example, consider a  capsule function that captures higher-order statistical moments as follows (for simplicity, we omit the mean and standard deviation),
\begin{equation}\label{eq:capsule}
\begin{split}
f_i(x_0,...,x_k) &  = \frac{1}{|\Na(i)|} \begin{bmatrix}
	\sum\limits_{k \in \Na(i)} a_{ik} x_k  \\
	\sum\limits_{k \in \Na(i)} a_{ik} x^2_k   \\
	\vdots \\
		\sum\limits_{k \in \Na(i)} a_{ik} x^p_k   \\
	\end{bmatrix}\\
\end{split}
\end{equation}

Figure~\ref{fig:capsule} shows an instance of applying  our graph capsule function   on a specific node. Consequently,  for an input feature matrix $\mathbf{X} \in \mathbb{R}^{N \times d} $, our graph capsule network will produce an output  $f(\mathbf{X}, \mathbf{L}) \in \mathbb{R}^{N \times h \times p} $ where $p$ is the number of instantiation parameters.

\noindent \textbf{Managing Graph Capsule Vector Dimension}: In the first layer, our graph capsule network receives an input $\mathbf{X} \in \mathbb{R}^{N \times d} $ and produces a non-linear output $f^{(1)}(\mathbf{X}, \mathbf{L}) \in \mathbb{R}^{N \times h_1 \times p} $. Since our graph capsule function produces a vector of $p$ dimension (for each input $d$ dimension), the feature  dimension of the output in subsequent layers can quickly blow up to an unmanageable value. To keep it in check, we restrict the feature  dimension of the output $f^{(\ell)}(\mathbf{X}, \mathbf{L})$ to be always $ \in \mathbb{R}^{N \times h_{\ell}\times p} $  at any middle $\ell^{th}-$layer of a GCAP-CNN (here $h_\ell$ represents the hidden dimension of that layer). This can be accomplished in two ways  1) either by flattening the last two dimension of $f(\mathbf{X}, \mathbf{L})$   and carrying out graph convolution in usual way (see Equation~\ref{eq:main} for an example)  2) or  by taking the weighted combination of $p-$dimension capsule vectors  (this is similar to performing attention mechanism) at each node as performed in~\cite{sabour2017dynamic}. We leave the second approach for our future work.  Thus in a nutshell,   our graph capsule network in  $\ell^{th}-$layer ($\ell>1$) receives an input $f^{(\ell-1)}(\mathbf{X}, \mathbf{L}) \in \mathbb{R}^{N \times h_{\ell-1}\times p} $ and produces an output  $f^{(\ell)}(\mathbf{X}, \mathbf{L}) \in \mathbb{R}^{N \times h_{\ell} \times p} $.

\noindent \textbf{Graph Capsule Function with Statistical Moments}: In this paper,  we consider higher-order statistical moments as instantiation parameters because they are permutationally invariant and can nicely be computed through   matrix-multiplication operations in a fast manner. To see exactly how, let $f_p(\mathbf{X}, \mathbf{L})$ be the output  matrix corresponding to $p^{th}$ dimension. Then, we can  compute  $f_p^{(\ell)}(\mathbf{X}, \mathbf{L})$ containing statistical moments as instantiation parameters as follows, 

\begin{equation}\label{eq:main}
\begin{split}
& f_p^{(\ell)}(\mathbf{X}, \mathbf{L})  \\
& = \sigma\Big(\sum_{k=0}^{K} \mathbf{L}^{k}(\underbrace{f^{(\ell-1)}_{F}(\mathbf{X}, \mathbf{L}) \odot ...\odot f^{(\ell-1)}_{F}(\mathbf{X}, \mathbf{L})}_{p \text{ times}})\mathbf{W}_{pk}^{(\ell)} \Big) \\
\end{split}
\end{equation}

 where $\odot$ is a hadamard  product. Here to keep the feature dimensions in check from growing, we flatten the last two dimension of the input as  $f^{(\ell-1)}_{Flat}(\mathbf{X}, \mathbf{L}) \in \mathbb{R}^{N \times h_{\ell-1}p}$
 and performs usual graph convolution operation followed by a  linear transformation with $\mathbf{W}_{pk}^{(\ell)} \in \mathbb{R}^{h_{\ell-1} p\times h_{\ell}}$  as the learning weight parameter. Note that here $p$ is used to denote both the capsule dimension as well   the order of statistical moments.

\noindent \textbf{Graph Capsule Function with Polynomial Coefficients}: As mentioned earlier, the quality of instantiation parameters depend upon their capability to encode and decode the  input values. Therefore, we   seek capsule functions which are bijective in nature i.e., guaranteed to preserve everything about the local neighborhood. For instance, one consider coefficients of polynomial as instantiation parameters by taking the set of local node  feature values as roots, 

\vspace{-0.5em}
\begin{equation}\label{eq:capsule}
\begin{split}
f_i(\cdot) &  =  \frac{1}{|\Na(i)|} \begin{bmatrix}
\sum\limits_{k \in \Na(i)} x_k \\
\sum\limits_{k_1, k_2 \in \Na(i)} x_{k_1}x_{k_2}    \\
\sum\limits_{k_1, k_2,k_3 \in \Na(i)}x_{k_1} x_{k_2}x_{k_3} \\
\vdots \\
x_0 x_1 \dots x_{k-1} x_k  \\
\end{bmatrix}\\
\end{split}
\end{equation}
\vspace{-0.5em}

One can show that from a given full set of  polynomial coefficients, we are guaranteed to recover back all the original node values (upto permutation). However, the first issue with   this approach is that they are   expensive to compute at each node. Specifically,  a combinatorial algorithm without fast fourier transform takes $\BigO(k^2)$  complexity to compute where $k$ is the number of roots. Also, there is numerical instability issue associated with computing   polynomial coefficients. There are   ways to deal with    these kind issues but we leave pursuing this direction   for our future work. 

In short, our graph capsule idea is powerful  and can be employed in any type of GCNN model for either solving graph semi-supervised learning problem or performing sequence learning on graphs using Graph Recurrent Neural Network models (GCRNNs) or doing link prediction via Graph Autoencoders (GAEs) or/and for generating synthetic graphs through Graph Generative Adversarial models (GGANs).

\section{Designing Graph Permutation Invariant Layer}\label{sec:permutation_inv}

In this section, we focus on the second limitation of GCNN model regarding achieving permutation invariance for graph classification purpose. Before presenting  our novel invariant layer in   GCAPS-CNN model, we first discuss the shortcomings of Max-Sort Pooling Layer which is the next popular choice after aggregation for achieving invariance.

\subsection{Problems with Max-Sort Pooling Layer}\label{sec:max-issue}

We design a test to determine whether the invariant graph feature constructed by a model has any degree of certainty  to produce the same output for \emph{sub-graph isomers} or not. 

\noindent \textbf{Sub-Graph Isomorphism Feature Test}: Consider two graphs $G_1=(V_1,E_1)$ and $G_2=(V_2,E_2)$ such that $G_1$ is isomorphic to a sub-graph of $G_2$. Let $\mathbf{f}_1,\mathbf{f}_2 \in \mathbb{R}^k$   be the invariant feature vector (w.r.t. to graph isomorphism) of $G_1, G_2$ respectively. Then, we define sub-graph isomorphism feature test as a criteria providing guarantee that each elements of $\mathbf{f}_1$ and $\mathbf{f}_2$ are comparable under certain notion i.e., $\mathbf{f}_{1i} \equiv \mathbf{f}_{2i}$ for any $i \in [1, k]$. Here  $\equiv$  represents a comparison operator defined in a sensible way. Satisfying this test is  very desirable for graph classification problem since it is quite likely that sub-graph isomers of a graph belong to the same class label. This property helps the model to  learn  $w_i$ weight parameter appropriately which is shared     across the  same input place i.e., $\mathbf{f}_{1i}$ and  $\mathbf{f}_{2i}$.

\begin{prop}\label{prop:isomorphic-test} Let $\mathbf{f}_1,  \mathbf{f}_2\in \mathbb{R}^k$ be the feature vectors containing top $k-\max$ node values in   sorted order for graphs $G_1, G_2$ respectively and given $G_1$ is sub-graph isomorphic to $G_2$. Then the Max-Sort Pooling Layer \emph{fails} the Sub-graph Isomorphism Feature Test owing to the comparison done with respect to node ordering.
\end{prop}

\noindent\textbf{Remarks}: Max-Sort Pooling layer  \emph{fails} the test because it does not guarantee that  $\mathbf{f}_{1i} \not\equiv \mathbf{f}_{2i}$ for any $i \in [1, k]$. Here  $ \not\equiv$ (not comparable) operator represents that the node corresponding to values $\mathbf{f}_{1i}$ and $\mathbf{f}_{2i}$  may not be the same in sub-graph isomers. Even including a single node (value) in $\mathbf{f}_2$ vector which is not present in $G_1$ can mess up the whole comparision order of  $\mathbf{f}_1$ and $\mathbf{f}_2$ elements.  As a result, in  Max-Sort Pooling layer the comparison is not always guaranteed to be sensible  which makes the problem of  learning  weight parameters harder. In general, any  invariant graph feature vector that relies on node ordering will fail this test.

\subsection{Covariance as Permutation Invariant Layer}
Our novel idea of   permutation invariant features in GCAPS-CNN model is  computing the covariance of $f(\mathbf{X}, \mathbf{L})$  layer output  given as follows,

\begin{equation}\label{eq:cov}
\begin{split}
\Ca(f(\mathbf{X}, \mathbf{L})) & =  \frac{1}{N}(f(\mathbf{X}, \mathbf{L})-\mu)^{T} (f(\mathbf{X}, \mathbf{L})-\mu)  \\
\end{split}
\end{equation}

Here $\mu$ is the mean of  $f(\mathbf{X}, \mathbf{L}) $ output and $\Ca(\cdot)$ is a   covariance function. Since covariance function is differentiable and does not depends upon the order of row elements,  it can serve as a permutation invariant layer in  GCAPS-CNN model. Also, it is fast in computation due to a single matrix-multiplication operation. Note that we flatten the last two dimension of GCAPS-CNN layer output   $f(\mathbf{X}, \mathbf{L}) \in \mathbb{R}^{N \times h\times p}  $ in order to compute the   covariance.

Moreover, covariance provides much richer information about the data by including  shapes,  norms and angles (between node hidden features) information rather  than just providing the mean of  data. Infact in multivariate normal distribution, it is  used as a  statistical parameter to approximate the normal density and thus also reflects information about the data distribution. This particular property along with invariance has   been exploited before in~\cite{kondor2003kernel} for computing similarity between two set of vectors. One can also think about   fitting multivariate normal distribution on $f(\mathbf{X}, \mathbf{L})$   but it involves computing inverse of  covariance matrix which is  computationally expensive. 
 
Since each element of covariance matrix is invariant to node orders, we can flatten the symmetric covariance matrix $C\in \mathbb{R}^{hp \times hp}$ to construct the graph invariant feature vector $\mathbf{f} \in \mathbb{R}^{(hp+1)hp/2} $. On an  another positive note, here the output dimension of $\mathbf{f}$ does not depend upon $N$ number of nodes and can be adjusted according to computational constraints.

\begin{prop}\label{prop:cov-isomorphic-test} 
Let $\mathbf{f}_1,  \mathbf{f}_2\in \mathbb{R}^k$ be the feature vectors containing covariance elements of  node feature matrices for graphs $G_1, G_2$ respectively and given $G_1$ is sub-graph isomorphic to $G_2$. Then the covariance invariant layer \emph{pass} the Sub-Graph Isomorphism Feature Test  owing to the comparison done with respect to feature dimensions.
\end{prop}
\noindent\textbf{{Remarks}}: It is quite straightforward to see that the feature dimension order of a node  does not depend upon the graph node ordering and hence   the order is same across all    graphs. As a result, each elements of $\mathbf{f}_1$ and $\mathbf{f}_2$ are always comparable. To be more specific, covariance output compares both the norms sand angles between the corresponding pairs of feature dimension vectors in two graphs.

\section{Designing GCAP-CNN with Global Features}\label{sec:global_features}

Besides guaranteeing permutation invariance in GCAP-CNN model, another important desired characteristic of  graph classification model is to capture global structure (or features) of a graph.  For instance, considering only node degree (as a node feature) is a local  information and not much helpful towards solving graph classification problem. On the other hand, considering spectral embedding as a node feature  takes global  piece of information into account and have been proven successful  in serving as a node vector for problems dealing with graph semi-supervised learning.  
We define global features that takes full graph structure into account during   their computation. While local features only depend upon some (at-most) $k-$hop node  neighbors.

Unfortunately, the basic design of GCNN model   can   only capture local structure information of the graph at each node. We make this loose statement more concrete with the following theorem.

\begin{theorem}\label{thm:k_hop_info}:
	\textit{Let  $G$ be a graph with $\mathbf{L}\in \mathbb{R}^{N \times N}$ graph Laplacian and $\mathbf{X}\in \mathbb{R}^{N \times d}$ node feature matrix. Let $f^{(\ell)}(\mathbf{X}, \mathbf{L})$ be the output function of a $\ell^{th}$  GCNN layer equipped with polynomial filters of degree $k$. Then $[f^{(\ell)}(\mathbf{X}, \mathbf{L})]_{i}$ output at $i^{th}$ node (i.e., $i^{th}$ row in $f^{(\ell)}(\cdot)$) depends upon ``only" on the   input values of  neighbors distant at most ``$k\ell-$hops" away.}
\end{theorem}

\noindent\textbf{Proof}:  We can   proof this statement by mathematical induction. It is easy to see that the base case  $\ell=1$ holds true. Lets assume it also holds true for $f^{(\ell-1)}(\mathbf{X}, \mathbf{L})$ i.e., $i^{th}$ node output   depends upon neighbors distant upto $k\times (\ell-1)$ hop away. Then in  $f^{(\ell)}(\mathbf{X}, \mathbf{L})   = \sigma \Big( g\big(f^{(\ell-1)}(\mathbf{X}, \mathbf{L}), \mathbf{L}\big) \mathbf{W}^{(\ell)}\Big)$  we focus on the term,

\begin{equation}\label{eq:proof}
\begin{split}
g(\mathbf{X}, \mathbf{L}) & =[f^{(\ell-1)}(\mathbf{X}, \mathbf{L}), \dots, \mathbf{L}^{k}f^{(\ell-1)}(\mathbf{X}, \mathbf{L})] \\
\end{split}
\end{equation}

 particularly the last term involving $\mathbf{L}^{k}f^{(\ell-1)}(\mathbf{X}, \mathbf{L})$. Matrix multiplication of $\mathbf{L}^{k}$ with $f^{(\ell-1)}(\mathbf{X}, \mathbf{L})$   will result in $i^{th}$ node to include all node information which are at-most $k-$hop distance away. But since a node in $f^{(\ell-1)}(\mathbf{X}, \mathbf{L})$ at a distance $k-$hops (from $i^{th}$ node) can contain information upto $k\times (\ell-1)$ hops, we have $i^{th}$ node containing information at-most $k+k(\ell-1)=k\ell$ hops distance away. 

\noindent \textbf{Remarks}: Above theorem~\ref{thm:k_hop_info} establishes that  GCNN model with $\ell$ layers can capture only $k\ell-$hop local-hood structure information at each node. Thus, employing GCNN for graph classification with say aggregation layer can capture only   average variation of $k\ell-$hop local-hood information over the whole graph. To include more  global information   about the graph one can either increase $k$ (i.e, choose higher order graph convolution filters) or $\ell$ (i.e, the depth of GCNN model). Both these choices increases model complexity and thus would require more data samples to reach satisfying results. However among the two, we prefer increasing the depth of GCNN model because the first choice leads to increase in the breadth of the GCNN layer (see footnote~\ref{fnm:1} about $g(\mathbf{X}, \mathbf{L})$ in Section~\ref{sec:model}) and based on the current understanding of deep learning theory, increasing the depth is   favored more over  the breadth.

For cases  where graph node features are missing, it is a common practice to take node degree as a node feature. Such practices can work for problems like graph semi-supervised where local-structure information drives node output labels (or classes). But in graph classification global features governs the output labels and hence taking node degree is not sufficient. Of course, we can go for a very deep GCNN model that will allows us to exploit more global information but requires higher sample  complexity to achieve satisfying results. 

To balance the two (model complexity with depth vs. required sample complexity), we propose to incorporate \textsc{Fgsd} features in our GCAP-CNN model   computed  at each node. As shown in~\cite{verma2017hunt} \textsc{Fgsd} features capture global information about the graph and can also be computed in fast manner. Specifically, at each  $i^{th}$ node   \textsc{Fgsd} features are computed  as the histogram of the multi-set formed by taking the harmonic distance between all   nodes and the $i^{th}$ node. It is given by, 

\begin{equation}{\label{eq:spectral}}
\Sa(x,y)=\sum_{n=0}^{N-1} \frac{1}{\lambda_n} (\phi_n(x)-\phi_n(y))^2   \\
\end{equation}
where $\Sa(x,y)$ is the harmonic distance,  $x,y$ are any graph nodes and  ${\lambda_n},\phi_n(\cdot)$ is the $n^{th}$ eigenvalue and eigenvector respectively. 

In our experiments, we employ   these features only for datasets  where node feature are missing (specifically for social network datasets in our case). Although this strategy can   always be used by concatenating  \textsc{Fgsd} features with original node feature values to capture more global information. Further  inspired from Weisfeiler-lehman graph kernel~\cite{shervashidze2011weisfeiler} which also concatenate features in each labeling iteration, we also propose to pass concatenated   outputs from intermediate layers to our covariance   and  fully connected layers. Finally, our whole end-to-end GCAP-CNN learning model     is guaranteed   to produce the same output  for isomorphic graphs. 
       

\section{Experiment and Results}\label{sec:exp_results}

\renewcommand{\arraystretch}{2}
\begin{table*}[t!]
	\centering
	\fontsize{7}{8}\selectfont

	\begin{tabular}{ @{} >{\raggedright}p{2cm} |    K{1.6cm}  !{\vrule width0.8pt} K{1.6cm}  !{\vrule width0.8pt} K{1.6cm} !{\vrule width0.8pt} K{1.6cm}   !{\vrule width0.8pt} K{2cm}   !{\vrule width0.8pt}K{1.6cm}   !{\vrule width0.8pt} K{1.6cm} | }
		
		\multirow{1}{*}{\textbf{Dataset}} &      \multicolumn{1}{c!{\vrule width0.8pt}}{PTC} &	\multicolumn{1}{c!{\vrule width0.8pt}}{PROTEINS}  &  \multicolumn{1}{c!{\vrule width0.8pt}}{NCI1} &  \multicolumn{1}{c!{\vrule width0.8pt}}{NCI109} &	 \multicolumn{1}{c!{\vrule width0.8pt}}{D \& D} &	 \multicolumn{1}{c!{\vrule width0.8pt}}{ENZYMES} \\
		
		
		{\textbf{(No. Graphs)}} &     {$344$} &	 {$1113$}  &  {$4110$} & 	 {$4127$} &	 {$1178$} &	 {$600$} \\  
		
		\multirow{1}{*}{\textbf{(Max. Graph Size)}} &    {$109$} &	 {$620$}  &  {$111$} & 	 {$111$} &	 {$5748$} &	 {$126$} \\  
		
		\multirow{1}{*}{\textbf{(Avg. Graph Size)}}    &  {$25.56$} &
		{$39.06$}  &  {$29.80$} & 	 {$29.60$} &	 {$284.32$} &	 {$32.60$} \\  \Xhline{2\arrayrulewidth}
	\end{tabular}	
	
	\begin{center}
		Deep Learning Methods  
	\end{center}

	\begin{tabular}{ @{} >{\raggedright}p{1.94cm} |     K{1.6cm}  !{\vrule width0.8pt} K{1.6cm}  !{\vrule width0.8pt} K{1.6cm} !{\vrule width0.8pt} K{1.6cm}   !{\vrule width0.8pt} K{2cm}   !{\vrule width0.8pt}K{1.6cm}   !{\vrule width0.8pt} K{1.6cm} | }
		\hline
			DCNN[\citeyear{atwood2016diffusion}]      &$56.60 \pm 2.89$&  $61.29 \pm 1.60 $   &$56.61 \pm 1.04$&  $57.47 \pm 1.22$ &$58.09\pm 0.53$&  $42.44 \pm 1.76$  \\  \hline
		PSCN[\citeyear{niepert2016learning}]      &$62.29 \pm 5.68$&  $75.00 \pm 2.51$  &$76.34 \pm 1.68$ &  --- &    --- &   ---  \\  \hline

		ECC[\citeyear{simonovsky2017dynamic}]       & --- &  ---  &$76.82$&  $75.03 $ &$72.54$&  $45.67$   \\  \hline
		
		DGCNN[\citeyear{zhang2018end}]       &$58.59 \pm 2.47$&  $75.54 \pm 0.94 $   &$74.44\pm0.47$&   $75.03 \pm 1.72$ &$\mathbf{79.37\pm0.94}$&   $51.00 \pm 7.29$  \\  \hline
		\textbf{GCAPS-CNN}          &${ \mathbf{66.01\pm 5.91}}$&  $\mathbf{76.40 \pm 4.17}$    &${ \mathbf{82.72 \pm 2.38}}$&  ${ \mathbf{81.12 \pm 1.28}}$  & $ {77.62 \pm 4.99}$ &  ${ \mathbf{61.83 \pm 5.39}}$   \\  \hline
	\end{tabular}

	\begin{center}
		Graph Kernels
	\end{center}
	
	\begin{tabular}{ @{} >{\raggedright}p{2cm} |    K{1.6cm}  !{\vrule width0.8pt} K{1.6cm}  !{\vrule width0.8pt} K{1.6cm} !{\vrule width0.8pt} K{1.6cm}   !{\vrule width0.8pt} K{2cm}   !{\vrule width0.8pt}K{1.6cm}   !{\vrule width0.8pt} K{1.6cm} | }	 
		\hline
		RW[\citeyear{gartner2003graph}]       &$57.85 \pm 1.30$&  $74.22 \pm 0.42$   &$>1$ Day&  $>1$ Day &$>1$ Day&  $24.16 \pm 1.64$   \\  \hline
		SP[\citeyear{borgwardt2005shortest}]       &$58.24 \pm 2.44$&  $75.07 \pm 0.54$   &$73.00\pm0.24$&  $73.00 \pm 0.21$ &$>1$Day  &  $40.10 \pm 1.50$   \\  \hline
		GK[\citeyear{shervashidze2009efficient}]       &$57.26 \pm 1.41  $&  $71.67 \pm 0.55$   &$62.28 \pm 0.29 $&  $62.60 \pm 0.19$ &$78.45 \pm 1.11$&  $26.61 \pm 0.99$   \\  \hline
		WL [\citeyear{shervashidze2011weisfeiler}]    &$57.97 \pm 0.49$&  $74.68 \pm 0.49$   &$82.19 \pm 0.18$&  $\mathbf{82.46 \pm 0.24}$ &$\mathbf{79.78 \pm 0.36}$&  $52.22 \pm 1.26$   \\  \hline
		DGK[\citeyear{yanardag2015deep}]   &$60.08 \pm 2.55 $&  $75.68 \pm 0.54$   &$80.31 \pm 0.46$&  $80.32 \pm 0.33$ & $73.50 \pm 1.01$ &  $53.43 \pm 0.91$   \\  \hline
		MLG[\citeyear{kondor2016multiscale}]        &$63.26 \pm 1.48  $&  $76.34 \pm 0.72 $   &$81.75 \pm 0.24$&  $81.31 \pm 0.22$ &$78.18 \pm 2.56 $&  $61.81 \pm 0.99  $   \\  \hline
		\textbf{GCAPS-CNN}        &${ \mathbf{66.01\pm 5.91}}$&  $\mathbf{76.40 \pm 4.17}$   &$\mathbf{82.72 \pm 2.38}$&  ${81.12 \pm 1.28}$  & $ {77.62 \pm 4.99}$ &  $\mathbf{61.83 \pm 5.39}$   \\  \hline
	\end{tabular}
	
	\caption{Classification  accuracy   on bioinformatics datasets. 
		Result in \textbf{bold} indicates the  best reported classification accuracy. Top half of the table compares results  with   various deep learning approaches while bottom half compares results with graph kernels. `$>1$ day' represents that the computation exceed more than $24hrs$. `OMR' is out of memory error.} 
		\label{table:bio_results}
	
\end{table*}

\noindent \textbf{GCAPS-CNN Model Configuration}: We build  $\ell$ layer GCAPS-CNN     with following configuration: $Input \rightarrow GC(h,p) \rightarrow \dots \rightarrow GC(h,p) \rightarrow [M,C(\cdot)]\rightarrow FC(h) \rightarrow FC(h) \rightarrow Softmax$. Here $GC(h,p)$ represents a Graph Capsule CNN layer with $h$ hidden dimensions and $p$ instantiation parameters. As mentioned earlier, we take the intermediate output of each $GC(h,p)$ layers and form a concatenated tensor    which is subsequently pass through $ [M,C(\cdot)]$ layer which computes mean and covariance of the input. Output of $ [M,C(\cdot)]$ layer   is then  passed to two fully connected $FC$ layers  with again $h$ output dimensions and finally connects to a  softmax layer for computing class probabilities. In between intermediate layers, we use batch normalization and  dropout  technique to prevent overfitting along with $L2$ norm regularization. We set $\ell\in \{2, 3, 4\}$ depending upon the dataset size (towards higher   for larger dataset) and $h\in \{32, 64, 128\}$ for setting hidden dimension. We restrict $p\in [1, 4]$ for computing higher-order statistical moments due to computational constraints. Further, we employ ADAM optimization technique with  initial learning rate chosen from the set $\{10^{-1},\dots, 10^{-7}\}$ with a   decaying factor of $0.1$ after every few   epochs. Batch size is set according to the given dataset size and memory requirements. Number of epochs are  chosen from the set $\{100, 200, 500, 1000\}$. All the above mentioned hyper-parameters are tuned based on the training loss. Average classification accuracy based on $10-$fold cross validation error is reported for each dataset.  Our GCAPS-CNN code and data will be made  available at Github\footnote{https://github.com/vermaMachineLearning/Graph-Capsule-CNN-Networks/}.

\noindent \textbf{Datasets}:  To evaluate  our  GCAPS-CNN model, we perform graph classification tasks on variety of benchmark datasets. In first round, we used $6$ bioinformatics   datasets namely:   PTC, PROTEINS, NCI1, NCI109, D\&D, and ENZYMES. In second round, we used $5$   social network datasets namely: COLLAB, IMDB-BINARY, IMDB-MULTI, REDDIT-BINARY and  REDDIT-MULTI-5K. D\&D dataset contains $691$ enzymes and $587$ non-enzymes proteins structures.  For other datasets  details can be found in~\cite{yanardag2015deep}. Also for each dataset number  of graphs, maximum   and average number of nodes   is  shown in the Table~\ref{table:bio_results} and Table~\ref{table:social_results}.


\renewcommand{\arraystretch}{2}
\begin{table*}[t!]
	\centering
	\fontsize{7}{8}\selectfont
	
	\begin{minipage}[t]{1\linewidth}
		\begin{center}
			\begin{tabular}{ @{} >{\raggedright}p{2cm} |  K{2cm} !{\vrule width0.8pt} K{2cm}  !{\vrule width0.8pt} K{2cm}  !{\vrule width0.8pt} K{2cm} !{\vrule width0.8pt} K{2cm}   !{\vrule width0.8pt} K{2cm}     | }
				
				\multirow{1}{*}{\textbf{Dataset}} & 	\multicolumn{1}{c!{\vrule width0.8pt}}{COLLAB}  &  \multicolumn{1}{c!{\vrule width0.8pt}}{IMDB-BINARY} &	\multicolumn{1}{c!{\vrule width0.8pt}}{IMDB-MULTI}  &  \multicolumn{1}{c!{\vrule width0.8pt}}{REDDIT-BINARY} &  \multicolumn{1}{c!{\vrule width0.8pt}}{REDDIT-MULTI}  \\
				
				
				\textbf{(No. Graphs)} &  {$5000$}  &  {$1000$} &	 {$1500$}  &  {$2000$} & 	 {$5000$}   \\
				
				\multirow{1}{*}{\textbf{(Max. Graph Size)}} &  \multicolumn{1}{c!{\vrule width0.8pt}}{$492$}  &  {$136$} &	 {$89$}  &  {$3783$} & 	 {$3783$}  \\   
				
				\multirow{1}{*}{\textbf{(Avg. Graph Size)}} &  \multicolumn{1}{c!{\vrule width0.8pt}}{$ 74.49 $}  &  {$ 19.77 $} &	 {$ 13.00 $}  &  {$ 429.61$} & 	 {$ 508.5 $}  \\  \Xhline{2\arrayrulewidth}
				
			\end{tabular}
			
			\begin{center}
				Deep Learning Methods   
			\end{center}

			\begin{tabular}{ @{} >{\raggedright}p{2cm} |  K{2cm} !{\vrule width0.8pt} K{2cm}  !{\vrule width0.8pt} K{2cm}  !{\vrule width0.8pt} K{2cm} !{\vrule width0.8pt} K{2cm}   !{\vrule width0.8pt} K{2cm}     | }
				\hline
						DCNN[\citeyear{atwood2016diffusion}]  &  $52.11 \pm 0.71$  &$49.06 \pm 1.37 $&  $33.49 \pm 1.42$   & OMR &  OMR  \\  \hline
				PSCN[\citeyear{niepert2016learning}]       &  $72.60\pm 2.15$  &$71.00 \pm 2.29 $&  $45.23\pm2.84$   &$86.30 \pm 1.58$&  $49.10 \pm 0.70$     \\  \hline
		
				DGCNN[\citeyear{zhang2018end}]       &  $73.76 \pm 0.49 $  &$70.03\pm 0.86$&  $47.83\pm0.85$   & $76.02 \pm 1.73$ &  $48.70 \pm 4.54$     \\  \hline
				{\textbf{{GCAPS-CNN}}  }    &  $\mathbf{77.71 \pm 2.51}$  &$\mathbf{71.69 \pm 3.40}$&  $\mathbf{48.50 \pm 4.10}$   & $\mathbf{87.61 \pm 2.51}$&  $\mathbf{50.10 \pm 1.72}$ \\  \hline
				
			\end{tabular}
			
			\begin{center}
				Graph Kernels
			\end{center}
			\begin{tabular}{ @{} >{\raggedright}p{2cm} |  K{2cm} !{\vrule width0.8pt} K{2cm}  !{\vrule width0.8pt} K{2cm}  !{\vrule width0.8pt} K{2cm} !{\vrule width0.8pt} K{2cm}   !{\vrule width0.8pt} K{2cm}     | }
				\hline	
				GK[\citeyear{shervashidze2009efficient}]  &  $72.84 \pm 0.28 $  &$65.87 \pm 0.98$&  $43.89 \pm 0.38 $   &$77.34 \pm 0.18$&  $41.01 \pm 0.17$   \\  \hline
				DGK[\citeyear{yanardag2015deep}]   &  $73.09 \pm 0.25$  &$66.96 \pm0.56$&  $44.55\pm0.52$   &$78.04 \pm 0.39$&  $41.27 \pm 0.18$   \\  \hline
				{\textbf{{GCAPS-CNN}}  }    &  $\mathbf{77.71 \pm 2.51}$  &$\mathbf{71.69 \pm 3.40}$&  $\mathbf{48.50 \pm 4.10}$   & $\mathbf{87.61 \pm 2.51}$&  $\mathbf{50.10 \pm 1.72}$ \\  \hline
			\end{tabular}

		\end{center}
	\end{minipage}
	
	\caption{Classification  accuracy   on social network datasets.	Result in \textbf{bold} indicates the  best reported classification accuracy. Top half of the table compares results  with   various deep learning approaches while bottom half compares results with graph kernels. `$>1$ day' represents that the computation exceed more than $24hrs$. `OMR' is out of memory error.} 
	\label{table:social_results}
	
\end{table*}

\noindent \textbf{Experimental Set-up}: All experiments were performed on a single machine loaded with recently launched $2\times$NVIDIA TITAN VOLTA GPUs and $64$ GB RAM. 
We compare our method  with both  deep learning models and graph kernels.

\noindent \textbf{Deep Learning Baselines}: For deep learning approaches, we adopted $4$ recently proposed state-of-art graph convolutional neural networks namely: 
PATCHY-SAN (PSCN)~\cite{niepert2016learning}, Diffusion CNNs (DCNN)~[\cite{atwood2016diffusion}], Dynamic Edge CNN (ECC)~\cite{simonovsky2017dynamic}  and Deep Graph CNN (DGCNN)~\cite{zhang2018end}.  

\noindent \textbf{Graph Kernel Baselines}: We adopted $6$ state-of-art  graphs kernels for comparison namely: Random Walk (RW)~\cite{gartner2003graph}, Shortest Path Kernel (SP)~\cite{borgwardt2005shortest}, Graphlet Kernel (GK)~\cite{shervashidze2009efficient}, Weisfeiler-Lehman Sub-tree Kernel (WL)~\cite{shervashidze2011weisfeiler}, Deep Graph Kernels (DGK)~\cite{yanardag2015deep} and  Multiscale Laplacian Graph Kernels (MLK)~\cite{kondor2016multiscale}.

\noindent \textbf{Baselines Settings}: We adopted the same procedure from previous works~\cite{niepert2016learning,yanardag2015deep,zhang2018end} to make a fair comparison and used $10$-fold cross validation with  
LIBSVM~\cite{chang2011libsvm} library to report the classification performance for graph kernels. Parameters of SVM are independently tuned using  training folds data and   best average classification accuracies are reported   for each    method. 
For Random-Walk (RW) kernel, decay factor is chosen from $\{10^{-6},10^{-5}...,10^{-1}\}$. For Weisfeiler-Lehman (WL) kernel, we chose height of subtree kernel from $h\in\{2,3,4\}$.  For graphlet kernel (GK), we chose graphlets size $\{3,5,7\}$ and for deep graph kernels (DGK), 
we report the best classification accuracy obtained among: deep graphlet kernel, deep shortest path kernel and deep Weisfeiler-Lehman kernel. For Multiscale Laplacian Graph (MLG) kernel, we chose $\eta$ and $\gamma$ parameter of the algorithm from $\{0.01,0.1,1\}$, radius size from $\{1,2,3,4\}$, and level number from $\{1,2,3,4\}$. For  diffusion-convolutional neural networks (DCNN), we chose number of hops from $\{2,5\}$. 
For the rest, best reported results were borrowed from papers PATCHY-SAN  ($k=10$)~\cite{niepert2016learning}, ECC~\cite{simonovsky2017dynamic} (without edge labels since all other methods also relies on only node labels) and DGCNN (with sorting layer)~\cite{zhang2018end}, since the experimental setup was the same and a fair comparison can be made. In short, we follow    the same procedure as mentioned in previous papers. Note: some results  are not present because either they are not previously reported  or source code not available to run them.

\noindent \textbf{Graph Classification Results}: From Table~\ref{table:bio_results}, it is   clear that our GCAPS-CNN  model \textbf{consistently outperforms}  most of the considered deep learning methods on       bioinformatics datasets (except on   D\&D dataset)   with a significant margin of $\mathbf{1\textbf{\%}-6\textbf{\%}}$ classification accuracy gain (highest   being on NCI1 dataset). 

Again, this trend is continued to be the same on   social network datasets as shown in Table~\ref{table:social_results}.  Here, we were able to achieve upto  $\mathbf{4\textbf{\%}}$ accuracy gain on COLLAB dataset and rest were around $\mathbf{1\textbf{\%}}$ gain   with \textbf{consistency} when compared against other deep learning approaches.

Our GCAPS-CNN is also very competitive with state-of-art graph kernel methods. It again  show a consistent performance gain of   $\mathbf{1\textbf{\%}-3\textbf{\%}}$ accuracy  (highest being on PTC dataset)  on many bioinformatic datasets when compared against with strong graph kernels.  While other considered deep learning methods are   not even close enough to  beat graph kernels on many of these  datasets. It is worth mentioning that the most deep learning models (like ours) are also scalable     while graph kernels are more fine tuned towards handling small graphs.  

For social network datasets, we have a significant gain of atleast $\mathbf{4\textbf{\%}-9\textbf{\%}}$ accuracy (highest being on REDDIT-MULTI dataset) against graph kernels as observed in  Table~\ref{table:social_results}. But this is expected as deep learning methods tend to do better with the large amount of data available for training on social networks datasets. Altogether, our GCAPS-CNN model shows very promising results against both  the current state-of-art  deep learning methods and graph kernels.

\section{Conclusion \& Future Work }

In this paper, we present a novel    Graph Capsule Network (GCAPS-CNN)  model based on the fundamental capsule idea to address some of the basic weaknesses of existing GCNN models. Our graph capsule network model by design   captures more local structure information than traditional GCNN and can   provide much richer representation of individual graph nodes or for the whole graph. For our purpose, we employ a capsule function that preserves statistical moments formation  since they are faster to compute. 

Furthermore, we propose a novel permutation invariant layer based on computing covariance in our GCAPS-CNN architecture to deal with graph classification problem which most GCNN models find  challenging. This covariance can again be computed in a fast manner and has   shown to be better than adopting aggregation or max-sort pooling layer. On the top, we also propose to equip our GCAPS-CNN model with  \textsc{Fgsd} features explicitly  to capture more global information in absence of node features. This is essential   to consider since non-deep GCNN models are not capable enough to exploit global information implicitly.  Finally, we show   GCAPS-CNN   superior performance on many bioinformatics and social network datasets in comparison with  existing   deep learning methods as well as strong graph kernels   and set the current state-of-the-art.

Our general idea of graph capsule is quite rich and can taken to another level by designing  more sophisticated capsule functions that are capable of preserving more information in a local pool. In our future work, we will investigate  various other capsule functions such as  polynomial coefficients (as instantiation parameters)  which comes with theoretical guarantees. Another choice, we will investigate is performing  kernel density estimation technique  in end-to-end deep learning framework  and understanding their   theoretical significance. Lastly, we will also explore the other approach of managing the graph capsule vector dimension as discussed in~\cite{sabour2017dynamic}.

\section*{Acknowledgement}
The research was supported in part by  US DoD DTRA grants HDTRA1-09-1-0050
and HDTRA1-14-1-0040, ARO MURI Award W911NF-12-1-0385 and NSF grants CNS 1618339
and CNS 1617729. 

\bibliography{refs}
\bibliographystyle{icml2018}

\end{document}